\documentclass[10pt,twocolumn,letterpaper]{article}

\usepackage[pagenumbers]{cvpr} 

\definecolor{cvprblue}{rgb}{0.21,0.49,0.74}
\usepackage[pagebackref,breaklinks,colorlinks,allcolors=cvprblue]{hyperref}

\def\confName{CVPR}
\def\confYear{2026}

\usepackage{amssymb}
\usepackage{booktabs}
\usepackage{multirow}
\usepackage{bm}
\usepackage{bbding}
\usepackage{wrapfig} 
\usepackage{xcolor} 
\usepackage{soul}
\usepackage{colortbl}

\usepackage{tablefootnote}
\usepackage{footnotehyper}

\usepackage{tcolorbox}
\usepackage{multirow}
\usepackage{graphicx}
\usepackage{wrapfig}
\usepackage{caption}
\newcommand{\VarSty}[1]{\textnormal{\ttfamily\color{blue!90!black}#1}\unskip}
\newcommand{\VarGSty}[1]{\textnormal{\ttfamily\color{green!50!black}#1}\unskip}

\usepackage{marvosym} 
\usepackage{amsmath}

\newcommand{\ourMethod}{DrivePI}

\title{{\ourMethod}: Spatial-aware 4D MLLM for Unified Autonomous Driving Understanding, 
Perception, Prediction and Planning
}

\author{
    Zhe Liu$^{1}$,\ 
    Runhui Huang$^{1}$,\
    Rui Yang$^{1}$,\ 
    Siming Yan$^{2}$, \\
        \vspace{5pt}
    Zining Wang$^{2}$,\
    Lu Hou$^{2}$,\
    Di Lin$^{3}$,\
    Xiang Bai$^{4}$,\
    Hengshuang Zhao$^{1, \text{\Letter}}$  \\ 
    $^{1}$The University of Hong Kong, \ 
    $^{2}$Yinwang Intelligent Technology Co. Ltd., \\
    $^{3}$Tianjin University, \ $^{4}$Huazhong University of Science and Technology \\
}

\begin{document}
\maketitle

\renewcommand\thefootnote{\text{\Letter}}
\footnotetext{Corresponding author}

\begin{abstract}

Although multi-modal large language models~(MLLMs) have shown strong capabilities across diverse domains, their application in generating fine-grained 3D perception and prediction outputs in autonomous driving remains underexplored.
In this paper, we propose {\ourMethod}, a novel spatial-aware 4D MLLM that serves as a unified Vision-Language-Action~(VLA) framework that is also compatible with vision-action~(VA) models. Our method jointly performs spatial understanding, 3D perception (\textit{i.e.}, 3D occupancy), prediction (\textit{i.e.}, occupancy flow), and planning (\textit{i.e.}, action outputs) in parallel through end-to-end optimization. To obtain both precise geometric information and rich visual appearance, our approach integrates point clouds, multi-view images, and language instructions within a unified MLLM architecture. We further develop a data engine to generate text-occupancy and text-flow QA pairs for 4D spatial understanding.
Remarkably, with only a 0.5B Qwen2.5 model as MLLM backbone, {\ourMethod} as a single unified model matches or exceeds both existing VLA models and specialized VA models.
Specifically, compared to VLA models, {\ourMethod} outperforms OpenDriveVLA-7B 
by 2.5\% mean accuracy on nuScenes-QA and reduces collision rate by 70\% over ORION~(from 0.37\% to 0.11\%) on nuScenes. Against specialized VA models, {\ourMethod} surpasses FB-OCC 
by 10.3 RayIoU for 3D occupancy on OpenOcc, reduces the mAVE from 0.591 to 0.509 for occupancy flow on OpenOcc, and achieves 32\% lower L2 error than VAD~(from 0.72m to 0.49m) for planning on nuScenes.
Code will be available at \url{https://github.com/happinesslz/DrivePI}.

\vspace{-10pt}
\end{abstract}

\section{Introduction}
\label{sec:intro}

\begin{figure}[t!]
\centering
\includegraphics[width=0.99\linewidth]{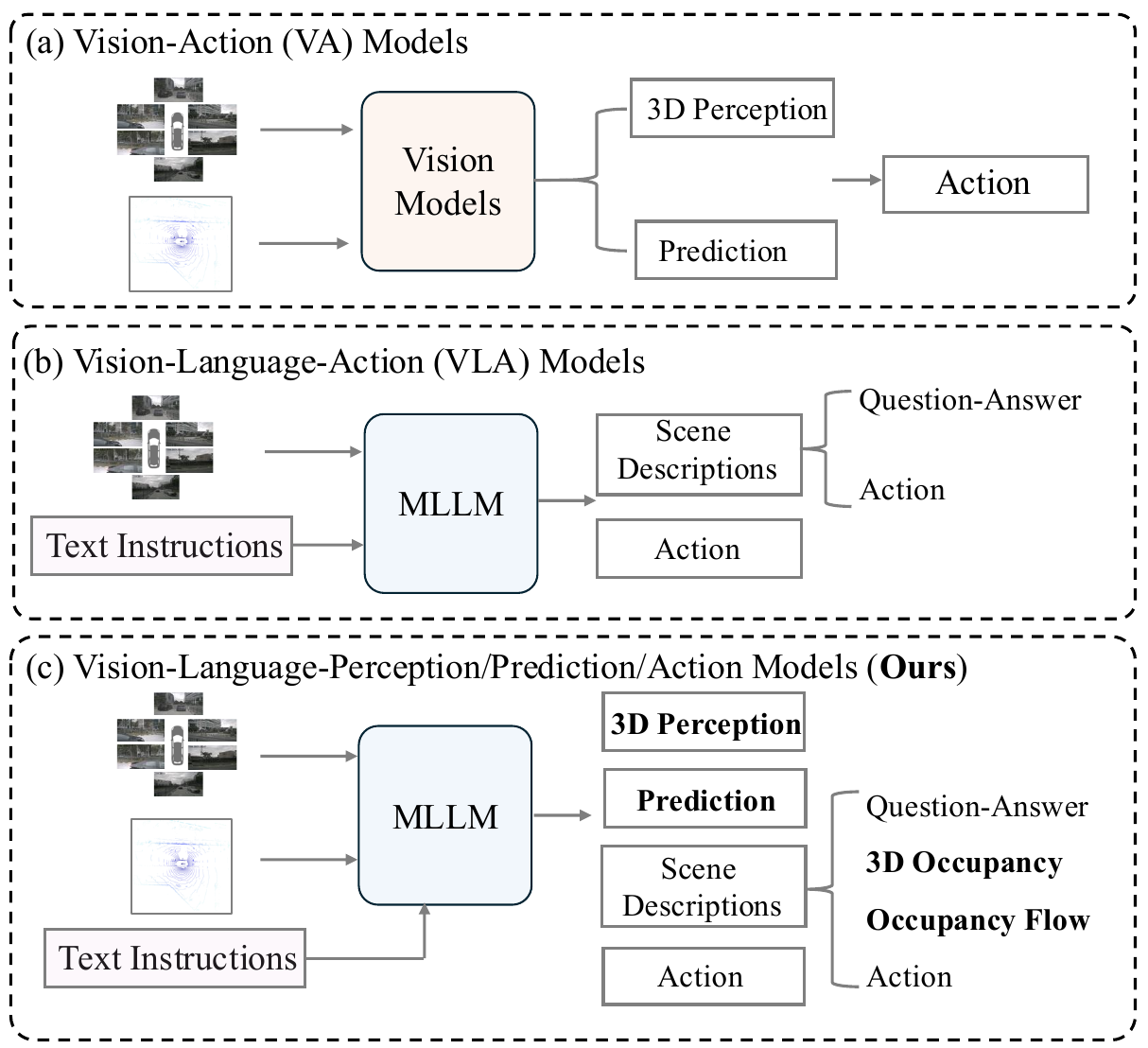}
\vspace{-5pt}
\caption{
(a) presents the pipeline of mainstream vision-action~(VA) models for end-to-end autonomous driving. (b) illustrates mainstream Vision-Language-Action~(VLA) models. (c) shows our  {\ourMethod}, which combines coarse-grained linguistic understanding with fine-grained 3D perception and prediction, inheriting advantages both existing VA models and VLA models.
}
\label{fig_intro}
\vspace{-15pt}
\end{figure}

In end-to-end autonomous driving systems, Vision-Action~(VA) models~\cite{hu2023planning,jiang2023vad} take vision information~(LiDAR point clouds, images) as inputs and output action signals, achieving substantial progress. Specifically, as shown in Figure~\ref{fig_intro}~(a), UniAD~\cite{hu2023planning} and VAD~\cite{jiang2023vad} adopt a modular framework that progresses from 3D perception to prediction, subsequently aggregating this information to generate final driving actions. Furthermore, FusionAD~\cite{ye2023fusionad} combines LiDAR point clouds and camera images to further enhance UniAD's performance.
Although these methods achieve promising results through their accurate spatial perception capabilities and the modular designs, they are limited in language-based scene interaction, which reduces user-friendliness.

To address these limitations, researchers~\cite{fu2025orion,zhou2025opendrivevla,wang2025omnidrive} have explored leveraging the powerful reasoning and human-like decision-making capabilities of MLLMs. Specifically, OpenDriveVLA~\cite{zhou2025opendrivevla} and ORION~\cite{fu2025orion} adopt a Vision-Language-Action~(VLA) framework that takes multi-view images and language instructions as inputs and generates actions, as illustrated in Figure~\ref{fig_intro}~(b). These VLA-based methods achieve superior interaction capabilities with scenarios and demonstrate enhanced user engagement. However, these VLA-based approaches usually struggle to guarantee reliable outputs due to the absence of fine-grained intermediate 3D perception and prediction outputs compared to the modular-design of VA models, consequently compromising interpretability and safety assurances.

Therefore, a natural question arises: \textit{Can we develop a unified framework that combines the precise spatial perception of VA models with the natural language interaction of VLA-based approaches?}
In this paper, we propose {\ourMethod}, a novel spatial-aware 4D MLLM that serves as a unified Vision-Language-Action~(VLA) framework for autonomous driving, as illustrated in Figure~\ref{fig_intro}(c). Here, we term it 4D MLLM as it outputs both 3D occupancy and flow, capturing fine-grained spatial-temporal dynamics. 
Unlike existing approaches that treat VA and VLA-based methods as separate paradigms, {\ourMethod} establishes a unified architecture that seamlessly integrates the spatial precision of VA models with the interpretability and interactive capabilities of VLA frameworks.

Specifically, {\ourMethod} exhibits four distinctive characteristics that differentiate it from existing approaches. first, unlike mainstream VLA-based methods that rely solely on camera images as input, {\ourMethod} introduces LiDAR as a complementary sensing modality, providing precise 3D geometric information that better elicits the spatial understanding capabilities of MLLMs. Second, {\ourMethod} generates intermediate fine-grained 3D perception~(\textit{e.g.}, 3D occupancy) and prediction~(\textit{e.g.}, occupancy flow) representations to ensure that the output features of the MLLM maintain reliable spatial perception capabilities, thereby enhancing both interpretability and safety guarantees for autonomous driving systems. Third, we develop enriched data engines that seamlessly integrate 3D occupancy and occupancy flow representations into natural language scene descriptions, enabling the model to reason about complex spatial-temporal dynamics through textual understanding. Fourth, {\ourMethod} serves as a unified model, which employs end-to-end joint optimization across all tasks including 3D perception, prediction, planning, and scene understanding.

In summary, our contributions are as follows: 
\begin{itemize}
    \item We propose {\ourMethod}, the first unified spatial-aware 4D MLLM framework that seamlessly integrates coarse-grained linguistic spatial understanding with fine-grained 3D perception capabilities, bridging the gap between VA-based and VLA-based paradigms in autonomous driving while inheriting the complementary strengths of both approaches;

    \item We incorporate LiDAR as a complementary sensing modality alongside camera imagery, providing high-precision 3D geometric information that better elicits the spatial understanding capabilities of MLLMs. Furthermore, {\ourMethod} enables accurate 3D perception (\textit{e.g.}, occupancy prediction) and prediction (\textit{e.g.}, occupancy flow), which effectively enhances the interpretability and safety assurances;

    \item  We develop three complementary spatial understanding benchmarks based on our data engine by constructing multiple question-answer~(QA) pairs: 3D occupancy perception for static scene understanding, occupancy flow prediction for dynamic motion analysis, and trajectory planning for decision-making evaluation. These benchmarks collectively assess different aspects of linguistic spatial reasoning capabilities across the temporal and spatial dimensions.
    
    \item Despite utilizing only a compact 0.5B parameter MLLM backbone, {\ourMethod} even outperforms  existing VA models in 3D occupancy and occupancy flow while maintaining comparable interactive capabilities with existing VLA frameworks in autonomous driving. 
\end{itemize}

\begin{figure*}[t!]
\vspace{-15pt}
\centering
\includegraphics[width=1.0 \linewidth]{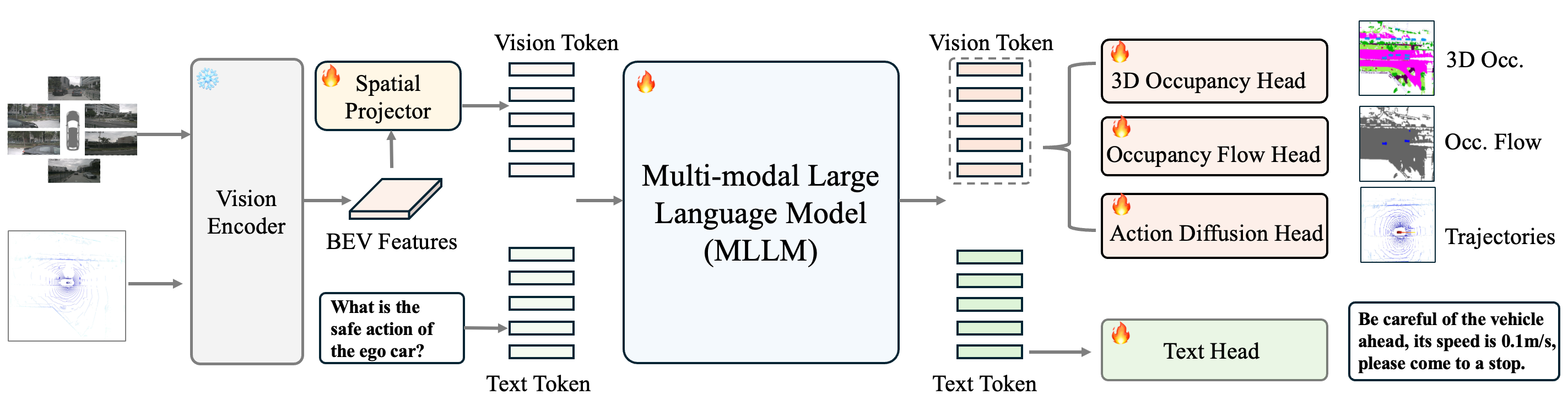}
\caption{
The pipeline of {\ourMethod} consists of the following steps. First, we employ a vision encoder to extract features from images and LiDAR data, obtaining latent BEV features that are then converted into vision tokens by a spatial projector. Next, we feed both vision tokens and text tokens into the MLLM to generate output tokens. The MLLM produces responses through four specialized heads: a text head for scene understanding in an auto-regressive manner, a 3D occupancy head for accurate spatial perception, an occupancy flow head for pixel-level motion prediction, and an action diffusion head for trajectory planning. 
}
\label{fig_framework}
\vspace{-10pt}
\end{figure*}

\section{Related Works}
\label{sec:formatting}

\noindent \textbf{Multimodal Large Language Model.}
With the rapid development of large language models~(LLMs)~\cite{bai2023qwen,team2024qwen2,yang2025qwen3,liu2024deepseek}, numerous works have attempted to incorporate additional modalities into LLMs, thereby expanding their application capabilities. Currently, multimodal large language models~(MLLMs) have been successfully adopted across various tasks, including image understanding~\cite{liu2023llava,liu2023improvedllava,bai2025qwen2,lu2024deepseekvl,chen2024internvl}, video understanding~\cite{lin2024vila,wang2022internvideo,wang2024internvideo2}, and semantic understanding~\cite{lai2024lisa,bai2024one,ren2024pixellm,wang2024git,zhang2024psalm,zheng2025villa}, demonstrating the enormous potential of LLMs for downstream applications. Consequently, an increasing number of works~\cite{wu2025spatial,ma2025spatialllm,mao2025spatiallm,zhou2025roborefer,yang2025thinking,abdolmaleki2025gemini} have begun exploring how to achieve spatial intelligence through MLLMs. Specifically, VSI-Bench~\cite{yang2025thinking} establishes a comprehensive benchmark that collects diverse indoor images and generates a series of questions designed to evaluate 3D spatial understanding, thereby validating the spatial reasoning capabilities of MLLMs. Gemini Robotics-ER~\cite{abdolmaleki2025gemini} predicts metric 3D bounding boxes from single images and further achieves open-vocabulary 3D object detection. Following this direction, Seed1.5-VL~\cite{guo2025seed1} enhances the perception capabilities of MLLMs and simultaneously processes 2D or 3D grounding tasks alongside other tasks (\textit{e.g.}, OCR, spatial understanding) from images using a unified MLLM architecture. However, these methods mainly achieve coarse-grained spatial perception such as describing relationships among objects or predicting object-level 3D bounding boxes.

\noindent \textbf{End-to-End Autonomous Driving.}
The autonomous driving system has undergone tremendous changes due to the development of the end-to-end model. Many works~\cite{jiang2023vad,hu2023planning,li2024ego,sun2024sparsedrive,ye2023fusionad,weng2024drive} have explored how to perform perception, prediction, and planning tasks in an end-to-end model, greatly improving the upper limit of planning performance and reducing the complexity of the autonomous driving system. UniAD~\cite{hu2023planning} adopts a modular architecture to process each task by different modules and jointly train these tasks in an end-to-end manner. VAD~\cite{jiang2023vad} proposes a vectorized representation and enables interaction among perception, mapping, and planning queries through different transformer decoders~\cite{vaswani2017attention}. Although these methods demonstrate promising planning results, they lack the ability to interact with users through natural language descriptions, resulting in reduced user-friendliness.

\noindent \textbf{Vision Language Action Model.}
Leveraging the powerful and versatile capabilities of MLLMs, several researches~\cite{xu2024drivegpt4,xu2025drivegpt4,shao2024lmdrive,sima2024drivelm,tian2024drivevlm,wang2023drivemlm,wang2025omnidrive} have successfully developed MLLMs for autonomous driving systems. 
These methods exploit the advanced reasoning capabilities of MLLMs to generate rich scene descriptions and high-level driving commands, thereby enhancing interpretability of end-to-end planning. Furthermore, to fully unleash comprehension abilities of MLLMs, recent works~\cite{zhou2025opendrivevla,li2025drivevla,jiang2025diffvla,fu2025orion,li2025recogdrive,hwang2024emma} integrate the planning module directly into MLLMs to achieve vision-language-action~(VLA) frameworks, enabling the integration of reasoning and planning tasks. For example, OpenDriveVLA~\cite{zhou2025opendrivevla} employs a Bird's-Eye-View~(BEV) representation to generate distinct tokens corresponding to agents, static maps, and scene context. These tokens are subsequently projected into a unified semantic space and processed by the MLLM to facilitate information interactions for generating trajectories by a planning decoder.
In this paper, we propose {\ourMethod}, a new VLA framework that seamlessly integrates both coarse-grained and fine-grained granularities. {\ourMethod} preserves the interpretability of textual representations for perception and planning reasoning while simultaneously employing fine-grained decoding mechanisms to generate precise 3D occupancy and trajectory predictions.

\section{Methods}

Multi-modal large language models have garnered increasing attention in autonomous driving due to their capabilities for user interaction and human-like decision-making. However, existing LLM-based methods struggle to directly output fine-grained perception results (\textit{e.g.}, 3D occupancy and occupancy flow) through next-token prediction, which are readily achievable by VA approaches. To address this limitation, we propose {\ourMethod}, a new VLA framework that achieves both coarse-grained linguistic understanding and fine-grained spatial perception, thereby inheriting the complementary advantages of both VA models and VLA frameworks. Next, we will introduce the details of {\ourMethod}.

\subsection{Overview}

As shown in Figure~\ref{fig_framework}, we present the pipeline of {\ourMethod}. First, to ensure that the inputs of MLLMs contain accurate geometric information, we incorporate LiDAR point clouds as additional inputs~(Note: LiDAR point clouds contains temporal information on the nuScenes dataset.), which provide precise 3D spatial information compared to camera images alone. This enhancement is instrumental in exploring the spatial perception capabilities of MLLMs. Second, we employ an advanced multi-modal vision encoder~\cite{liu2025unilion} to process multi-view images and LiDAR point clouds, subsequently converting them into a compact latent BEV feature representation.
Next, a spatial projector is utilized to map the latent BEV features into the language space and obtain vision tokens. These vision tokens, along with text tokens, are then fed into the MLLM. Finally, we utilize four specialized heads: a text head to generate responses for scene understanding in an auto-regressive manner, a 3D occupancy head for accurate 3D perception, an occupancy flow head for fine-grained motion prediction, and an action diffusion head for trajectory planning. Note that the MLLM is trainable, and all tasks are jointly optimized during training.

\begin{figure*}[t!]
\centering
\includegraphics[width=0.99\linewidth]{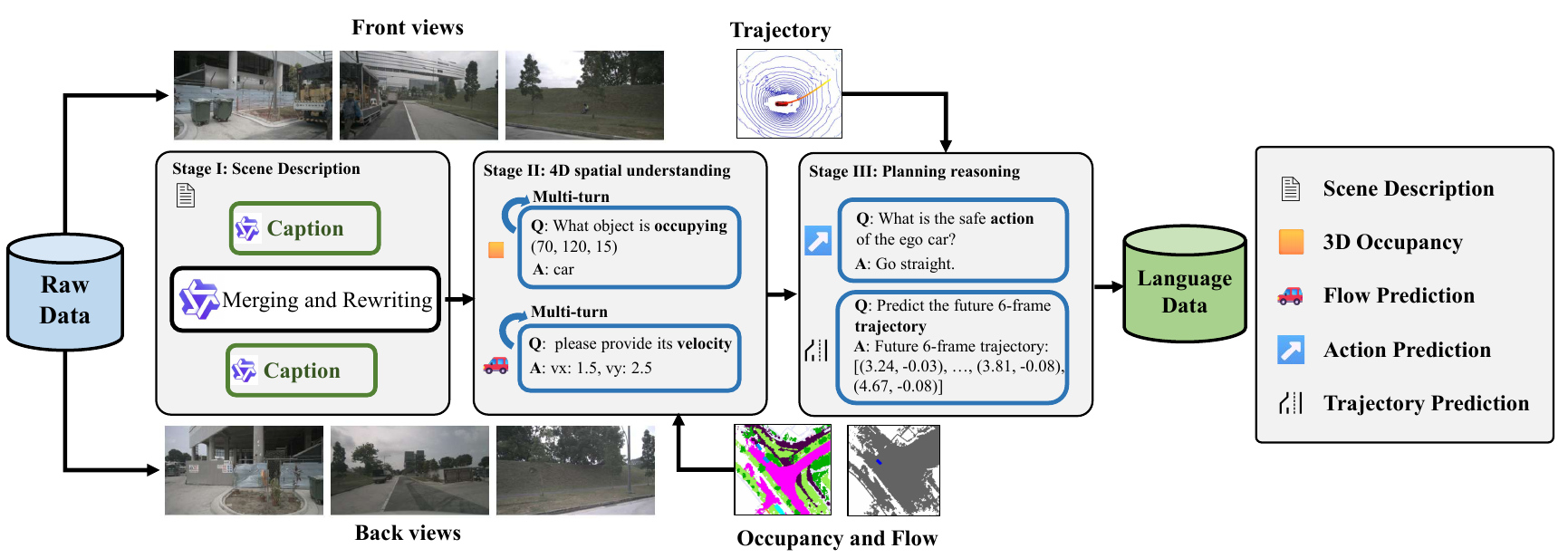}
\vspace{-8pt}
\caption{The illustration of our multi-stage data pipeline. We first generate captions of front and back views, respectively. Then, we use InternVL3-78B~(adopts Qwen2.5-72B~\cite{bai2025qwen2} as the language model) to combine these captions to merge and polish generated scene descriptions. Moreover, we generate text-occupancy and text-flow QA pairs based on occupancy and flow ground truth by multi-turn conversations to improve the 4D spatial understanding ability. Finally, we generate text-planning QA pairs to allow MLLM to predict the future actions of ego-vehicle.
}
\label{fig_data}
\vspace{-10pt}
\end{figure*}

\noindent\textbf{Spatial Projector.} Given the latent BEV feature $F_{bev} \in \mathbb{R}^{H\times W \times C}$, where $H \times W$ typically exceeds a resolution of $100 \times 100$, directly inputting these features at the pixel level into the MLLM would incur prohibitive computational costs. Here, $H, W, C$ are the height, width and channel dimensions of $F_{bev}$, respectively.
To address this challenge, we first patchify $F_{bev}$ into $N$ patches of size $K \times K$, producing visual features $F_{patch} \in \mathbb{R}^{N\times K^2 \times C}$, where $N=\frac{H}{K} \times \frac{W}{K}$. A conventional approach involves applying pooling operations to aggregate the $K \times K$ spatial features into a single representation, yielding pooled features $F_{pool} \in \mathbb{R}^{N \times 1 \times C}$. However, this approach typically results in the loss of fine-grained spatial information. 
Therefore, following~\cite{huang2025hires} for handling high-resolution image inputs in 2D MLLMs, we adopt a cross-attention mechanism with $F_{pool}$, $F_{patch}$, and $F_{patch}$ serving as query, key, and value, respectively. This design preserves more detailed spatial information. Finally, we employ a linear layer to transform the processed features to match the channel dimension $C_{l}$ of the MLLM's input hidden states, producing the final vision tokens $F_v \in \mathbb{R}^{N \times C_{l}}$.

\subsection{Coarse-grained Spatial Understanding}

In this paper, we define coarse-grained spatial understanding as the text-based descriptions that arises from the highly compressed nature of language compared to rich visual information (\textit{e.g.}, images, LiDAR point clouds). Although MLLMs demonstrate remarkable capabilities in many domains, exploring their spatial understanding remains a critical challenge, particularly in autonomous driving  where precise spatial reasoning is essential. Therefore, we develop three complementary spatial understanding benchmarks based on our data engine by constructing multiple question-answer~(QA) pairs: 3D occupancy perception for static scene understanding, occupancy flow prediction for dynamic motion analysis, and trajectory planning for decision-making evaluation. These benchmarks collectively assess different aspects of linguistic spatial reasoning capabilities across the temporal and spatial dimensions.

\noindent\textbf{Data Engine.} 
As shown in Figure~\ref{fig_data}, our pipeline consists of three main stages: caption annotation for scene understanding, 4D spatial understanding annotation, and planning reasoning annotation.
In the first stage, we employ InternVL3-78B~\cite{zhu2025internvl3} to generate scene descriptions for the front and back views separately. This strategy prevents potential confusion that may arise when MLLMs struggle to distinguish between different viewpoints. The captions from both views are then merged to construct a comprehensive description of the entire scene, which is further refined to ensure high descriptive quality.
The second stage aims to equip the MLLM with 4D spatial understanding capabilities. Specifically, we leverage ground-truth occupancy and flow data to generate diverse text-occupancy and text-flow QA pairs. These QA pairs focus on key tasks such as determining whether given positions are occupied, identifying corresponding object categories, and predicting velocity information. This enables {\ourMethod} to explore fine-grained 3D occupancy and flow occupancy in textual format compared to previous methods~\cite{xu2024drivegpt4,li2025recogdrive,shao2024lmdrive}.
In the final stage, we generate text-planning QA pairs to enhance planning interpretability based on future trajectory annotations of the ego-vehicle. These pairs require the MLLM to analyze the surrounding environment and provide high-level driving commands along with suggested trajectories.
This multi-stage pipeline ensures both versatility and high quality of the generated language dataset, enabling the MLLM to develop 4D spatial understanding and planning capabilities. 
For more details of data engine~(\textit{e.g.}, the design of prompts), please refer to our supplemental materials.

\subsection{Fine-grained Spatial Learning}

In this paper, we define fine-grained spatial learning as the integration of explicit spatial capabilities, including 3D occupancy, occupancy flow, and trajectory planning. While language descriptions effectively capture high-level semantic concepts and global spatial arrangements, they inherently lack the precision required for detailed spatial localization and geometric understanding essential in autonomous driving tasks. Therefore, we introduce fine-grained vision heads to address this.

\noindent\textbf{Fine-grained Visual Heads}.
We adopt three fine-grained visual heads to enable precise spatial capabilities: a 3D occupancy head for volumetric scene understanding, an occupancy flow head for temporal dynamics modeling, and an action diffusion head for trajectory planning. Specifically, to implement these fine-grained heads, we first extract the corresponding vision tokens $F_{v}^* \in \mathbb{R}^{N \times C_l}$ from the multi-modal representation. Subsequently, we employ a linear projection layer to transform $F_{v}^*$ into $F_{v}^{out} \in \mathbb{R}^{N \times K^2C}$. We then reshape $F_{v}^{out}$ into a spatial feature map $F_{out} \in \mathbb{R}^{H \times W \times C}$, where $H$ and $W$ denote the height and width of the feature map, respectively. Based on the spatially-organized features $F_{out}$, we can seamlessly integrate our three prediction heads following existing VA models. For more details of each head, please refer to the supplementary material.

\subsection{Loss Function}
We define the total loss as a weighted summation of four components: $L_{llm}$ for the text head, $L_{occ}$ for the occupancy head, $L_{flow}$ for the occupancy flow head, and $L_{action}$ for the action diffusion head. Therefore, the total loss $L_{total}$ is formulated as:
\begin{equation}
L_{total} =   \lambda_1 L_{llm} +\lambda_2 L_{occ} + \lambda_3 L_{flow} + \lambda_4 L_{action},
\end{equation}
where $\lambda_1,\lambda_2,\lambda_3,\lambda_4$ are the balancing weights for text scene understanding, 3D occupancy perception, occupancy flow prediction, and trajectory planning, respectively. All tasks are optimized jointly in an end-to-end manner.

\section{Experiments}

\begin{table*}[t!]
\caption{3D occupancy and occupancy flow performance on the OpenOcc validation set.
}
\vspace{-8pt}
\small
\setlength{\tabcolsep}{5pt}
\resizebox{1.0\linewidth}{!}{
\centering
\begin{tabular}{l|c|c|cc|ccc} 
\toprule
Method & VLM-based & OccScore$\uparrow$ & RayIoU~(3D Occ.)$\uparrow$ & mAVE~(Occ. Flow)$\downarrow$ & RayIoU$_{\mathrm{1m}}$ & RayIoU$_{\mathrm{2m}}$ & RayIoU$_{\mathrm{4m}}$ \\
\midrule
OccNeRF~\cite{zhang2023occnerf} & & 28.5 & 31.7 & \textbf{--} & 16.6 & 29.3 & 49.2 \\ 
RenderOcc~\cite{pan2024renderocc} & & 33.0 & 36.7 & \textbf{--} & 20.3 & 32.7 & 49.9\\
LetOccFlow~\cite{liu2025let} & & 36.4 & 40.5 & \textbf{--} & 25.5 & 39.7 & 56.3\\
OccNet~\cite{tong2023scene} & & 35.7 & 39.7 & \textbf{--} & 29.3 & 39.7 & 50.0\\
BEVDetOcc-SF~\cite{huang2022bevdet4d} & & 33.0 & 36.7 & 1.420 & 31.6 & 37.3 & 41.1\\
FB-Occ~\cite{li2023fb} & & 39.2 & 39.0 & 0.591 & 32.7 & 39.9 & 44.4\\
F-Occ~\cite{zhao20243d} & & 41.0 & 39.9 & 0.491 & 33.9 & 40.7 & 45.2\\
CascadeFlow~\cite{liao2024cascadeflow} & & 40.9 & 39.6 & 0.470 & 33.5 & 40.3 & 45.0 \\
ALOcc-Flow-3D~\cite{chen2025alocc} & & 43.0 & 41.9 & 0.556 & 35.6 & 42.8 & 47.4 \\
\midrule
\rowcolor[gray]{0.95}
{\ourMethod}~(Ours) &$\checkmark$ & \textbf{49.3} & \textbf{49.3} & \textbf{0.509} & \textbf{45.0}  & \textbf{50.0}  & \textbf{52.9} \\
\bottomrule
\end{tabular}
 }
\label{tab:nuscenes_openocc_val_results}
\end{table*}

\begin{table*}[t!]
\caption{Planning performance on the nuScenes validation set. Note that our unified model {\ourMethod} does not incorporate ego status during training by default to avoid potential shortcut learning.
}
\vspace{-8pt}
\small
\setlength{\tabcolsep}{10pt}
\resizebox{1.0\linewidth}{!}{
\centering
\begin{tabular}{l|c|c|cccc|cccc}
\toprule
& & & \multicolumn{4}{c|}{\emph{L2 (m)$\downarrow$}} & \multicolumn{4}{c}{\emph{Col. (\%)$\downarrow$}} \\
Method & VLM-based & Ego Status & 1s & 2s & 3s & avg. & 1s & 2s & 3s & avg. \\
\midrule
ST-P3~\cite{} & & & 1.33 & 2.11 & 2.90 & 2.11 & 0.23 & 0.62 & 1.27 & 0.71 \\
FF~\cite{hu2021safe} & & & 0.55 & 1.20 & 2.54 & 1.43 & 0.06 & 0.17 & 1.07 & 0.43 \\
EO~\cite{khurana2022differentiable} & & & 0.67 & 1.36 & 2.78 & 1.60 & 0.04 & 0.09 & 0.88 & 0.33 \\
UniAD~\cite{hu2023planning} & & & 0.48 & 0.96 & 1.65 & 1.03 & 0.05 & 0.17 & 0.71 & 0.31 \\
VAD~\cite{jiang2023vad} & & & 0.41 & 0.70 & 1.05 & 0.72 & 0.07 & 0.17 & 0.41 & 0.22 \\
VAD~\cite{jiang2023vad} & & $\checkmark$ & 0.17 & 0.34 & 0.60 & 0.37 & 0.07 & 0.10 & 0.24 & 0.14 \\
\midrule
OmniDrive~\cite{wang2025omnidrive} & $\checkmark$ & $\checkmark$ & 0.14 & 0.29 & 0.55 & 0.33 & 0.00 & 0.13 & 0.78 & 0.30 \\
ORION~\cite{fu2025orion} & $\checkmark$ & $\checkmark$ & 0.17 & 0.31 & 0.55 & 0.34 & 0.05 & 0.25 & 0.80 & 0.37 \\
OpenDriveVLA-7B~\cite{zhou2025opendrivevla} & $\checkmark$ & $\checkmark$ & 0.20 & 0.58 & 1.21 & 0.66 & 0.00 & 0.22 & 0.55 & 0.25 \\
\midrule
\rowcolor[gray]{0.95}
{\ourMethod}~(Ours) & $\checkmark$ &  & \textbf{0.24} & \textbf{0.46} & \textbf{0.78} & \textbf{0.49} & \textbf{0.38} & \textbf{0.27} & \textbf{0.48}  & \textbf{0.38} \\
\rowcolor[gray]{0.95}
{\ourMethod}~(Ours) & $\checkmark$ & $\checkmark$ & \textbf{0.19} & \textbf{0.36} & \textbf{0.64} & \textbf{0.40} & \textbf{0.00} & \textbf{0.05} & \textbf{0.28} & \textbf{0.11} \\
\bottomrule
\end{tabular}
 }
\label{tab:nuscenes_planning_val_results}
\vspace{-10pt}
\end{table*}

\subsection{Datasets and Evaluation Metrics}

In this paper, we conduct comprehensive experiments on the nuScenes~\cite{caesar2020nuscenes} dataset, a large-scale autonomous driving benchmark. The dataset comprises 750 training scenes, 150 validation scenes, and 150 test scenes. nuScenes provides synchronized multi-modal sensor data including LiDAR point clouds and multi-view images from 6 cameras. For occupancy prediction, we employ OpenOcc~\cite{tong2023scene} as the primary occupancy evaluation. Moreover, to make a comprehensive comparison with most 3D occupancy methods, we evaluate the reuslts on  Occ3D~\cite{tian2023occ3d}. 
For occupancy flow, we utilize the occupancy flow annotations provided by OpenOcc~\cite{tong2023scene} to evaluate the performance of {\ourMethod} in fine-grained motion prediction. For trajectory planning evaluation, we follow established protocols from prior works~\cite{zhou2025opendrivevla,wang2025omnidrive} and employ open-loop evaluation metrics on the nuScenes dataset. For text understanding, we conduct experiments on the nuScenes-QA~\cite{qian2024nuscenes} dataset containing 377k training question-answer~(QA) pairs, and employ our proposed data generation pipeline to generate 84k training scene descriptions, 560k QA pairs for 4D spatial reasoning and 24k for planning reasoning, culminating in a comprehensive training dataset of over 1.0M QA pairs.

For evaluation metrics, we adopt the established metrics to evaluate the performance of {\ourMethod}. For 3D occupancy and occupancy flow, we use RayIoU~\cite{liu2024fully} and the mean Average Velocity Error (mAVE)~\cite{caesar2020nuscenes}, respectively. For planning, we follow previous methods~\cite{sun2024sparsedrive,zhou2025opendrivevla} and use L2 distance error and collision rate metrics. To evaluate text understanding, we adopt the official evaluation metrics from the nuScenes-QA benchmark~\cite{qian2024nuscenes}. For 4D spatial understanding~(see Table~\ref{tab:ab_size}), we adopt the following metrics: 1)~accuracy for occupancy status, occupancy classification, action status, 2)~mAVE for occupancy flow, 3)~L2 distance error and collision rate for planning.


\subsection{Implementation Details}

To balance computational efficiency, we adopt Qwen2.5-0.5B~\cite{bai2025qwen2} as our base model. The training of {\ourMethod} comprises two stages. In the first stage, we freeze the vision encoder and the MLLM, then train the spatial projector to align visual representations with the language embedding space for 1 epoch using the generated captions of our data engine.
In the second stage, we jointly optimize the spatial projector, MLLM, and all task-specific heads~(\textit{i.e.}, text head, 3D occupancy head, occupancy flow head, and action diffusion head) for 1 epoch while keeping only the vision encoder frozen.
For loss balancing, we adopt the balancing weights of $\lambda_1=\lambda_2=\lambda_3=\lambda_4=1$. All experiments are conducted on 8$\times$NVIDIA L40S GPUs.

\subsection{Main Results}
We comprehensively evaluate {\ourMethod} across four key tasks: text understanding, 3D occupancy, occupancy flow, and trajectory planning. For 3D occupancy and occupancy flow evaluation, we conduct comparisons on the OpenOcc~\cite{tong2023scene} benchmark, which provides annotations for both 3D occupancy and flow on nuScenes. For trajectory planning, we directly adopt the annotations provided by the official nuScenes dataset. Moreover, to demonstrate the text understanding capability of {\ourMethod}, we report results on the nuScenes-QA benchmark, which offers diverse visual question-answering pairs tailored to autonomous driving scenarios.
Unless otherwise specified, all experiments are conducted using our unified 0.5B model architecture with shared parameter weights across all tasks.

\noindent\textbf{3D Ocuupancy and Occupancy Flow on OpenOcc.}
As shown in Table~\ref{tab:nuscenes_openocc_val_results}, we compare the 3D occupancy and occupancy flow performance of {\ourMethod} with existing approaches. {\ourMethod} achieves superior performance across multiple metrics, with an OccScore of 49.3\%, RayIoU of 49.3\%, and mAVE of 0.509. {\ourMethod} surpasses the representative method FB-OCC~\cite{li2023fb} 
by 10.3 RayIoU for 3D occupancy and reduces flow mAVE from 0.591 to 0.509.
Notably, {\ourMethod} outperforms the previous state-of-the-art method ALOcc-Flow-3D~\cite{chen2025alocc} by 6.3\% in OccScore, 7.4\% in RayIoU, and reduces mAVE by 0.047, establishing new state-of-the-art results with only a 0.5B MLLM backbone. This demonstrates the effectiveness of {\ourMethod} in achieving remarkable 4D fine-grained spatial perception capabilities within a VLA framework.

\noindent\textbf{Planning on nuScenes.}
To further evaluate the planning capabilities of {\ourMethod} as a VLA model, we conduct experiments on the nuScenes benchmark~\cite{caesar2020nuscenes}.  As shown in Table~\ref{tab:nuscenes_planning_val_results},  we compare the performance of our {\ourMethod} against both traditional end-to-end VA methods and VLM-based approaches. {\ourMethod} achieves an L2 error of 0.40 and a collision rate of 0.11\% with ego status, outperforming the end-to-end VA method VAD~\cite{jiang2023vad} and VLA method OpenDriveVLA-7B~\cite{zhou2025opendrivevla} by 0.03\% and 0.14\% in collision rate, respectively.  Notably, {\ourMethod} reduces collision rate by 70\% compared to the recent ORION~\cite{fu2025orion} 
(from 0.37\% to 0.11\%). Without ego status, {\ourMethod} achieves 32\% lower L2 error than VAD (from 0.72m to 0.49m). 
These results demonstrate the effectiveness of {\ourMethod} as a VLA model for planning tasks.

\noindent\textbf{Text Understanding on nuScenes-QA.}
Text understanding capability is crucial for VLA models, as it enables autonomous driving systems to interpret and reason based on natural language, thereby supporting more human-like decision-making. We validate the text understanding performance of {\ourMethod} on the nuScenes-QA~\cite{qian2024nuscenes} benchmark. As shown in Table~\ref{tab:nuscenes_text_val_results}, {\ourMethod} with only a 0.5B model size achieves 60.7\% accuracy, outperforming OpenDriveVLA-7B~\cite{zhou2025opendrivevla} by 2.5\%. These results demonstrate the promising text understanding capabilities of {\ourMethod}.

\begin{table*}[t!]
\caption{Text Understanding performance on the nuScenes-QA validation set. Ext., Cnt., Obj., Sts., Cmp. and Acc. are short for exist, count, object, status, comparison, and the overall accuracy.
}
\vspace{-8pt}
\small
\setlength{\tabcolsep}{20pt}
\resizebox{1.0\linewidth}{!}{
\centering
\begin{tabular}{l|c|c|c|c|c|c} 
\toprule
Method & Ext.$\uparrow$ & Cnt.$\uparrow$ & Obj.$\uparrow$ & Sts.$\uparrow$ & Cmp.$\uparrow$ & Acc.$\uparrow$\\
\midrule
LLaMA-AdapV2~\cite{gao2023llama} & 19.3 & 2.7 & 7.6 & 10.8 & 1.6 & 9.6 \\
LLaVA1.5~\cite{liu2023improvedllava} & 45.8 & 7.7 & 7.8 & 9.0 & 52.1 & 26.2 \\
LiDAR-LLM~\cite{yang2025lidar} & 74.5 & 15.0 & 37.8 & 45.9 & 57.8 & 48.6 \\
BEVDet+BUTD~\cite{qian2024nuscenes} & 83.7 & 20.9 & 48.8 & 52.0 & 67.7 & 57.0 \\
OpenDriveVLA-0.5B~\cite{zhou2025opendrivevla} & 83.9 & 22.0 & 50.2 & 57.0 & 68.4 & 58.4 \\
OpenDriveVLA-3B~\cite{zhou2025opendrivevla} & 84.0 & 22.3 & 50.3 & 56.9 & 68.5 & 58.5 \\
OpenDriveVLA-7B~\cite{zhou2025opendrivevla} & 84.2 & \textbf{22.7} & 49.6 & 54.5 & \textbf{68.8} & 58.2 \\
\midrule
\rowcolor[gray]{0.95}
{\ourMethod}~(Ours) & \textbf{85.3} & {22.4} & \textbf{57.5} & \textbf{59.1} & {68.3} & \textbf{60.7}\\
\bottomrule
\end{tabular}
 }
 \vspace{-5pt}
\label{tab:nuscenes_text_val_results}
\end{table*}

\begin{table}[t!]
\caption{3D occupancy performance on the Occ3D-nuScenes validation set. * indicates that {\ourMethod} is trained exclusively on the 3D occupancy task of Occ3D-nuScenes.
}
\vspace{-5pt}
\small
\setlength{\tabcolsep}{2pt}
\resizebox{1.0\linewidth}{!}{
\centering
\begin{tabular}{l|c|c|ccc} 
\toprule
Method & VLM-based & RayIoU$\uparrow$ & RayIoU$_{\mathrm{1m}}$ & RayIoU$_{\mathrm{2m}}$ & RayIoU$_{\mathrm{4m}}$ \\
\midrule
RenderOcc~\cite{pan2024renderocc} & & 19.5 & 13.4 & 19.6 & 25.5 \\
SimpleOcc~\cite{gan2023simple} & & 22.5 & 17.0 & 22.7 & 27.9 \\
BEVFormer~\cite{li2022bevformer} & & 32.4 & 26.1 & 32.9 & 38.0 \\
BEVDet-Occ~\cite{huang2021bevdet} & & 32.6 & 26.6 & 33.1 & 38.2 \\
FB-Occ~\cite{li2023fb} &  & 33.5 & 26.7 & 34.1 & 39.7 \\
SparseOcc~\cite{liu2024fully} &  & 36.1 & 30.2 & 36.8 & 41.2 \\
OPUS~\cite{wang2024opus} &  & 41.2 & 34.7 & 42.1 & 46.7 \\
\midrule
\rowcolor[gray]{0.95}
{\ourMethod}*~(Ours) & $\checkmark$ & \textbf{46.0} & \textbf{42.2} & \textbf{46.7} & \textbf{49.2} \\
\bottomrule
\end{tabular}
 }
\vspace{-5pt}
\label{tab:nuscenes_occ_val_results}
\end{table}

\begin{table}[t!]
\caption{Ablation study for text head and vision head in {\ourMethod}.}
\vspace{-8pt}
\small
\centering
\setlength{\tabcolsep}{2pt}
\resizebox{1.0\linewidth}{!}{
\begin{tabular}{c|c|c|c|c|cc|c}
\toprule
\multicolumn{1}{c|}{\multirow{2}{*}{\#}} & \multicolumn{1}{c|}{\multirow{2}{*}{Text Head}} & \multicolumn{1}{c|}{\multirow{2}{*}{Vision Head}} & \multicolumn{1}{c|}{\emph{3D Occ.}} & \multicolumn{1}{c|}{\emph{Occ. Flow}} & \multicolumn{2}{c|}{\emph{Planning}} & \multicolumn{1}{c}{\emph{QA}} \\
& & & RayIoU$\uparrow$ & mAVE$\downarrow$ & L2$\downarrow$ & Col.$\downarrow$ & Acc.$\uparrow$ \\
\midrule
\textit{I} & $\checkmark$  & \textbf{--}  &\textbf{--}  &-- &-- &-- &\textbf{61.2} \\
\textit{II} & \textbf{--}  & $\checkmark$ & 47.5 &0.69 &1.02 &0.39 &-- \\
\textit{III} & $\checkmark$  &  $\checkmark$ & \textbf{49.3} & \textbf{0.51} & \textbf{0.50} & \textbf{0.38} & {60.7} \\
\bottomrule
\end{tabular}
 }
 \vspace{-8pt}
\label{tab:ab_com}
\end{table}

\noindent\textbf{3D Occupancy on Occ3D.} Beyond the unified model {\ourMethod} trained jointly on all tasks, we also train {\ourMethod} exclusively on the 3D occupancy task of Occ3D to enable comprehensive comparisons with existing approaches on the Occ3D benchmark~\cite{tian2023occ3d}. 
As shown in Table~\ref{tab:nuscenes_occ_val_results}, {\ourMethod} achieves state-of-the-art performance with 46.0\% RayIoU, surpassing the previous best method OPUS~\cite{wang2024opus} by a significant margin of 4.8\%. Notably, {\ourMethod} is built on a MLLM architecture, which highlights its strong capabilities in fine-grained 3D perception, despite being primarily designed for multi-modal understanding.

\subsection{Ablation Study}
In this section, we conduct the ablation studies to illustrate the effectiveness of our designs in {\ourMethod}. Unless otherwise specified, we use the {\ourMethod}-0.5B by default.

\noindent\textbf{Ablation on Text Head and Vision Head.}
{\ourMethod} integrates text understanding, 3D perception, prediction, and planning within a unified VLA framework. For simplicity, we refer to the combination of the fine-grained 3D occupancy head, occupancy flow head, and trajectory planning head as the vision head.
To evaluate the impact of each component, we conduct ablation studies by removing individual heads, with quantitative results summarized in Table~\ref{tab:ab_com}. When only the text head is enabled (\textit{I}), {\ourMethod} achieves competitive performance with 61.2\% accuracy on the text understanding task. When only the vision head (\textit{i.e.}, occupancy, flow, and planning) is enabled (\textit{II}), {\ourMethod} still delivers promising results: 47.5\% RayIoU for 3D occupancy, 0.69 mAVE for occupancy flow, and 1.02 L2 error with 0.39 collision rate for trajectory planning~(we not use ego-vehicle information during training process).
When both text and vision heads are combined~(\textit{III}), {\ourMethod} achieves superior performance across most tasks. Compared to the vision-only setting (\textit{II}), the unified model~(\textit{III}) increases RayIoU by 1.8\%, reduces mAVE by 0.18, and decreases L2 error by 0.52. The main reason is that text understanding helps better align with the appropriate feature space for vision tasks. Furthermore, it maintains comparable text understanding performance with 60.7\% accuracy, close to the text-only setting (\textit{I}). These results verify the effectiveness of unifying text understanding with 3D perception, prediction, and planning tasks within a single VLA framework.

\begin{table*}[t!]
\caption{The ablation study of {\ourMethod} exploring data scaling. The columns of \emph{Occ. Status}, \emph{Occ. Class}, \emph{Occ. Flow}, \emph{Action Status} denote the occupancy status~(\textit{i.e.}, ``yes" or ``no"), occupancy category, the occupancy flow,  the action  commands~(\textit{i.e.}, ``straight", ``right", ``left", and ``stop"), respectively.
}
\vspace{-8pt}
\small
\centering
\setlength{\tabcolsep}{5pt}
\resizebox{1.0\linewidth}{!}{
\begin{tabular}{c|ccccc|ccc|c|c|c|c}
\toprule
 \multicolumn{1}{c|}{\multirow{2}{*}{Model Size}} & \multicolumn{5}{c|}{Training Data size} & \multicolumn{1}{c|}{\emph{Occ. Status}} & \multicolumn{1}{c|}{\emph{Occ. Class}}  & \multicolumn{1}{c|}{\emph{Occ. Flow}} & \multicolumn{1}{c|}{\emph{Action Status}} & \multicolumn{2}{c|}{\emph{Planning}}  & \multicolumn{1}{c}{\emph{QA}} \\
& Caption & OCC.  &Flow & Action  & QA & Acc.$\uparrow$ & Acc.$\uparrow$ &mAVE$\downarrow$ & Acc.$\uparrow$ & L2$\downarrow$ & Col.$\downarrow$ & Acc.$\uparrow$ \\
\midrule
0.5B & 84k  & 420k &140k & 24k  & 377k & 86.0 & 50.4  &0.91 & \textbf{83.8} & \textbf{0.79} & 0.63 & 60.7  \\
\midrule
\multirow{5}{*}{3B} & 84k & 28k  &--  & --  &-- & 73.0 & 14.3 & -- & -- & -- & --  \\
& 84k  & 56k  & -- &-- & --  & 74.2 & 22.4  &--& -- & -- & -- & --  \\
& 84k  & 280k   & -- &--& -- & 86.8 & 54.7   &--& -- & -- & -- & --  \\
& 84k  & 560k   & --&-- & -- & 87.0 & 59.2   &--& -- & -- & -- & --  \\
& 84k  & 420k &140k & 24k  & 377k & \textbf{88.6} & \textbf{59.8}   &\textbf{0.69} & 83.3 & 0.86 & \textbf{0.62} & \textbf{63.0}  \\
\bottomrule
\end{tabular}
}
\label{tab:ab_size}
\end{table*}

\begin{figure*}[t!]
\centering
\includegraphics[width=0.99\linewidth]{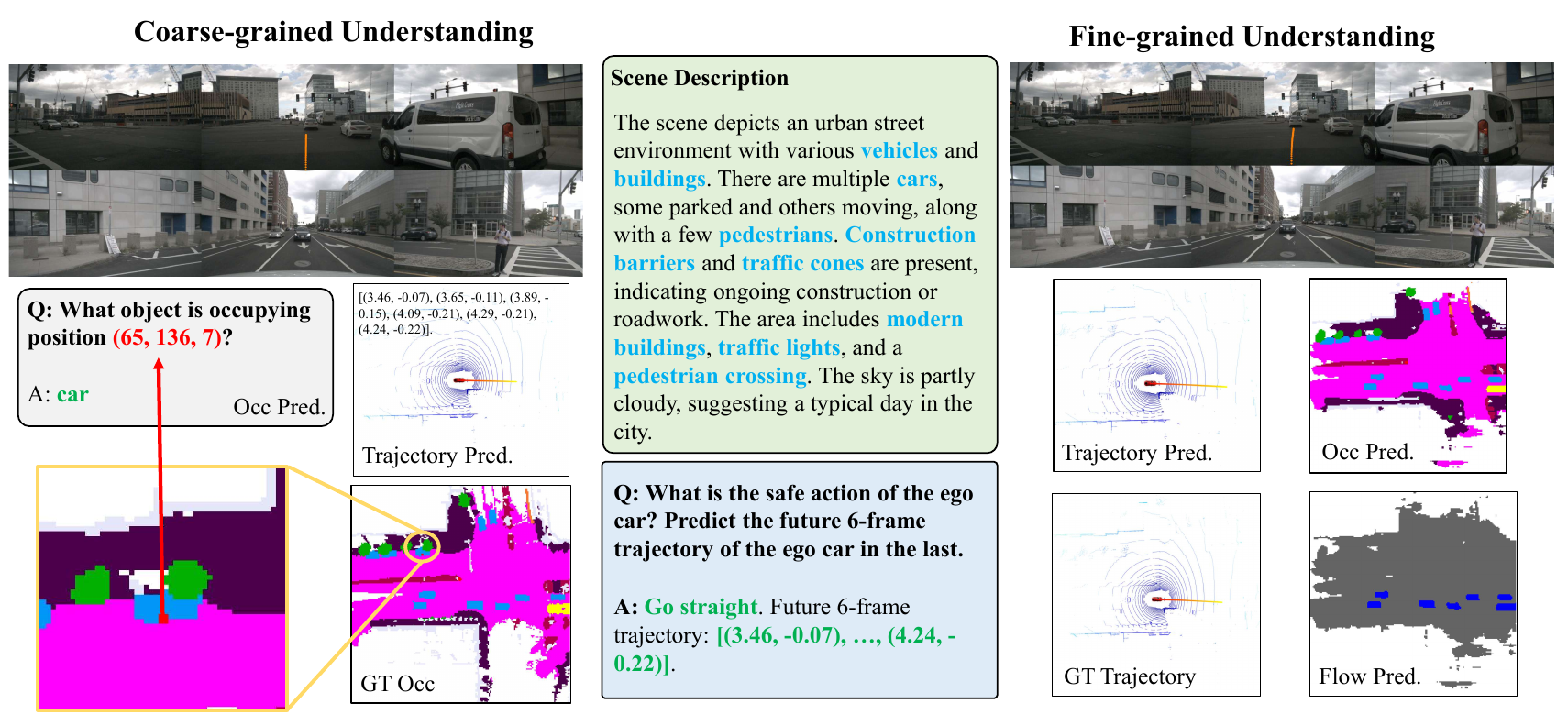}
\vspace{-5pt}
\caption{The visualization of coarse-grained and fine-grained understanding of {\ourMethod}. We present the results of scene description, 3D occupancy, trajectory prediction results to illustrate the coarse-grained understanding of {\ourMethod}. 
}
\vspace{-10pt}
\label{fig_vis}
\end{figure*}

\noindent \textbf{Scaling Text Data.} 
To explore the impact of different sizes of text data, we conduct a series of ablation studies by varying the amount of instruction-tuning data across multiple tasks, including predicting occupancy status, occupancy category, action status, and trajectory planning from our data engine, as well as QA from nuScenes-QA. As summarized in Table~\ref{tab:ab_size}, we mainly conduct the experiments on the Qwen-2.5 3B model using only the text head. 
When trained with only 112K samples (84K captions + 28K occupancy QA pairs), {\ourMethod} achieves only 73\% accuracy in occupied status prediction and 14.3\% accuracy in occupancy category prediction. When scaling the occupancy QA pairs from 28K to 560K, there is an obvious performance improvement with 14\% accuracy on predicting occupancy status and 44.9\% accuracy on predicting occupancy category. When further adding QA pairs for occupancy flow, action, and the official nuScenes-QA, the accuracy of predicting occupancy status and category brings performance improvements of 1.6\% and 0.6\%, respectively. Besides, the corresponding performance for occupancy flow, action status, and planning is 0.69 mAVE, 83.3\% Acc., and 0.62 collision rate. Finally, we also evaluate the 0.5B model with joint training of text and vision heads. We find that our 0.5B model even achieves better performance on action status prediction and L2 error despite having lower QA accuracy, demonstrating the effectiveness of {\ourMethod} in unifying text understanding, 3D perception, prediction, and planning.

\subsection{Visualization}
In this section, we present qualitative visualizations to showcase the multi-granularity understanding capabilities of {\ourMethod}. As illustrated in Figure~\ref{fig_vis}, we present the visualization of {\ourMethod} on scene descriptions, 3D occupancy, occupancy flow, action, and trajectory predictions. 
At a coarse-grained level, {\ourMethod} generates detailed scene descriptions and reasonable answers based on different instructions. While at a fine-grained level, it produces fine-grained results~(\textit{e.g.}, 3D occupancy, occupancy flow, and trajectory planning) through specialized decoding heads, enabling more precise prediction results. 
Besides, we have two important observations. 
First, regarding scene descriptions, {\ourMethod} demonstrates the ability to understand detailed appearance information (such as \textit{``the sky is partly cloudy''}), indicating that the vision encoder effectively preserves critical visual information even when transforming front-view images into Bird's-Eye-View representation.
Second, we observe strong alignment between coarse-grained and fine-grained predictions. For instance, the object category described linguistically at a coarse level corresponds to the fine-grained 3D occupancy prediction at the grid coordinate $(65, 136, 7)$. Similarly, the predicted trajectory exhibits similar consistency. This alignment across different levels of understanding validates the effectiveness of {\ourMethod} in unifying coarse-grained linguistic spatial understanding with fine-grained 3D perception capabilities, which enhances the interpretability and explainable decision-making of autonomous driving systems.

\section{Conclusion}
In this paper, we have presented a novel VLA framework, {\ourMethod}, which achieves both coarse-grained spatial understanding in text formats and fine-grained spatial perception comparable to VA models, thereby inheriting the advantages of both existing VA models and VLA frameworks. Notably, our {\ourMethod} employs only a 0.5B parameter LLM as its backbone, demonstrating remarkable efficiency. Despite utilizing only a compact 0.5B parameter MLLM backbone, {\ourMethod} with promising text understanding, even outperforms  existing VA models in 3D occupancy and occupancy flow while maintaining comparable interactive capabilities with existing VLA-based frameworks in autonomous driving. Finally, we hope this new VLA framework can inspire future research to enhance autonomous driving systems with improved interpretability and explainable decision-making through language reasoning and fine-grained 3D outputs. 

\noindent\textbf{Limitations.} Our approach still has two main limitations that warrant future investigation. First, we use a simple multi-task learning strategy to balance loss weights across tasks, which may not be optimal to obtain the best trade-off between competing objectives. Second, we do not incorporate reinforcement learning techniques that could enhance reasoning capabilities through trial-and-error learning, 
particularly for complex planning scenarios.

{
    \small
    \bibliographystyle{ieeenat_fullname}
    \bibliography{main}
}

\clearpage
\setcounter{page}{1}
\maketitlesupplementary

In this supplementary material, we present more ablation studies, the details of coarse-grained spatial understanding, the details of fine-grained spatial learning, and more visualizations in Sections ~\ref{sec:more_as}, ~\ref{sec:details_cos}, ~\ref{sec:details_fine} and ~\ref{sec:more_vis}, respectively.

\section{More Ablation Studies}
\label{sec:more_as}

\begin{table}[h!]
\vspace{-10pt}
\caption{Ablation study for the balancing weights in {\ourMethod}. }
\small
\centering
\setlength{\tabcolsep}{2pt}
\resizebox{1.0\linewidth}{!}{
\begin{tabular}{c|c|c|c|c|cc|c}
\toprule
\multicolumn{1}{c|}{\multirow{2}{*}{\#}} & \multicolumn{1}{c|}{\multirow{2}{*}{\emph{Occ. Weight}}} & \multicolumn{1}{c|}{\multirow{2}{*}{\emph{Flow Weight}}} & \multicolumn{1}{c|}{\emph{3D Occ.}} & \multicolumn{1}{c|}{\emph{Occ. Flow}} & \multicolumn{2}{c|}{\emph{Planning}} & \multicolumn{1}{c}{\emph{QA}} \\
& & & RayIoU$\uparrow$ & mAVE$\downarrow$ & L2$\downarrow$ & Col.$\downarrow$ & Acc.$\uparrow$ \\
\midrule
\textit{I} & 0.2  & 0.2  &48.1  &0.57 &0.46 &0.19 &\textbf{61.1}  \\
\textit{II} & 0.5  & 0.5 &\textbf{49.3}  &0.54  &0.49  &0.40  &60.9 \\
\textit{III} & 1.0  &  1.0 & \textbf{49.3} & \textbf{0.51} & \textbf{0.50} & \textbf{0.38} & {60.7} \\
\bottomrule
\end{tabular}
 }
 \vspace{-5pt}
\label{tab:ab_weight}
\end{table}

\begin{table}[t!]
\centering
\small
\caption{The learned importance weights of all hidden states in the MLLM  with Qwen-2.5 0.5B model, including the input embedding~(indexed as 0). The \textit{\textbf{Index}} and \textbf{Weight} column indicates the index and the learned importance weight of each hidden state. }
\vspace{-5pt}
\label{tab:layer_weights}
\setlength{\tabcolsep}{8pt}
\resizebox{1.0\linewidth}{!}{
\begin{tabular}{cccccc}
\toprule
\textbf{\textit{Index}} & \textbf{Weight} & \textbf{\textit{Index}} & \textbf{Weight} & \textbf{\textit{Index}} & \textbf{Weight} \\
\midrule
\textit{0} & 0.0328 & \textit{10} & 0.0375 & \textit{20} & 0.0463 \\
\textit{1} & 0.0332 & \textit{11} & 0.0381 & \textit{21} & 0.0466 \\
\textit{2} & 0.0337 & \textit{12} & 0.0388 & \textit{22} & 0.0472 \\
\textit{3} & 0.0341 & \textit{13} & 0.0397 & \textit{23} & 0.0477 \\
\textit{4} & 0.0346 & \textit{14} & 0.0409 & \textit{24} & 0.0468 \\
\textit{5} & 0.0350 & \textit{15} & 0.0422 &  &  \\
\textit{6} & 0.0355 & \textit{16} & 0.0435 &  &  \\
\textit{7} & 0.0360 & \textit{17} & 0.0449 &  &  \\
\textit{8} & 0.0365 & \textit{18} & 0.0455 &  &  \\
\textit{9} & 0.0370 & \textit{19} & 0.0458 &  &  \\
\bottomrule
\end{tabular}
}
\vspace{-5pt}
\end{table}

\begin{table}[h!]
\caption{The learned importance weights of all hidden states in the MLLM  with Qwen-2.5 3B model, including the input embedding~(indexed as 0). The \textit{\textbf{Index}} and \textbf{Weight} column indicates the index and the learned importance weight of each hidden state.}
\vspace{-5pt}
\centering
\setlength{\tabcolsep}{7pt}
\resizebox{1.0\linewidth}{!}{
\begin{tabular}{ccc|ccc|ccc}
\toprule
\textbf{\textit{Index}} & \textbf{Weight} & & \textbf{\textit{Index}} & \textbf{Weight} & & \textbf{\textit{Index}} & {\textbf{Weight}} \\
\midrule
\textit{0} & 0.0254 & & \textit{13} & 0.0259 & & \textit{26} & 0.0277 \\
\textit{1} & 0.0251 & & \textit{14} & 0.0260 & & \textit{27} & 0.0283 \\
\textit{2} & 0.0251 & & \textit{15} & 0.0261 & & \textit{28} & 0.0287 \\
\textit{3} & 0.0252 & & \textit{16} & 0.0262 & & \textit{29} & 0.0294 \\
\textit{4} & 0.0253 & & \textit{17} & 0.0263 & & \textit{30} & 0.0297 \\
\textit{5} & 0.0252 & & \textit{18} & 0.0264 & & \textit{31} & 0.0303 \\
\textit{6} & 0.0253 & & \textit{19} & 0.0265 & & \textit{32} & 0.0304 \\
\textit{7} & 0.0254 & & \textit{20} & 0.0266 & & \textit{33} & 0.0304 \\
\textit{8} & 0.0255 & & \textit{21} & 0.0271 & & \textit{34} & 0.0302 \\
\textit{9} & 0.0255 & & \textit{22} & 0.0265 & & \textit{35} & 0.0303 \\
\textit{10} & 0.0256 & & \textit{23} & 0.0251 & & \textit{36} & 0.0331 \\
\textit{11} & 0.0257 & & \textit{24} & 0.0258 & & & \\
\textit{12} & 0.0258 & & \textit{25} & 0.0267 & & & \\
\bottomrule
\end{tabular}
}
\vspace{-10pt}
\label{tab:3blayer_weights}
\end{table}

\begin{table*}[t!]
\caption{An example generated by our multi-stage data engine.}
\centering
\vspace{-2mm}
\begin{minipage}{0.85\textwidth}  
\centering
\begin{tcolorbox} 
\centering
\scalebox{0.9}{
\begin{tabular}{p{1.1\textwidth} c}  
\centering{\includegraphics[height=4.3cm]{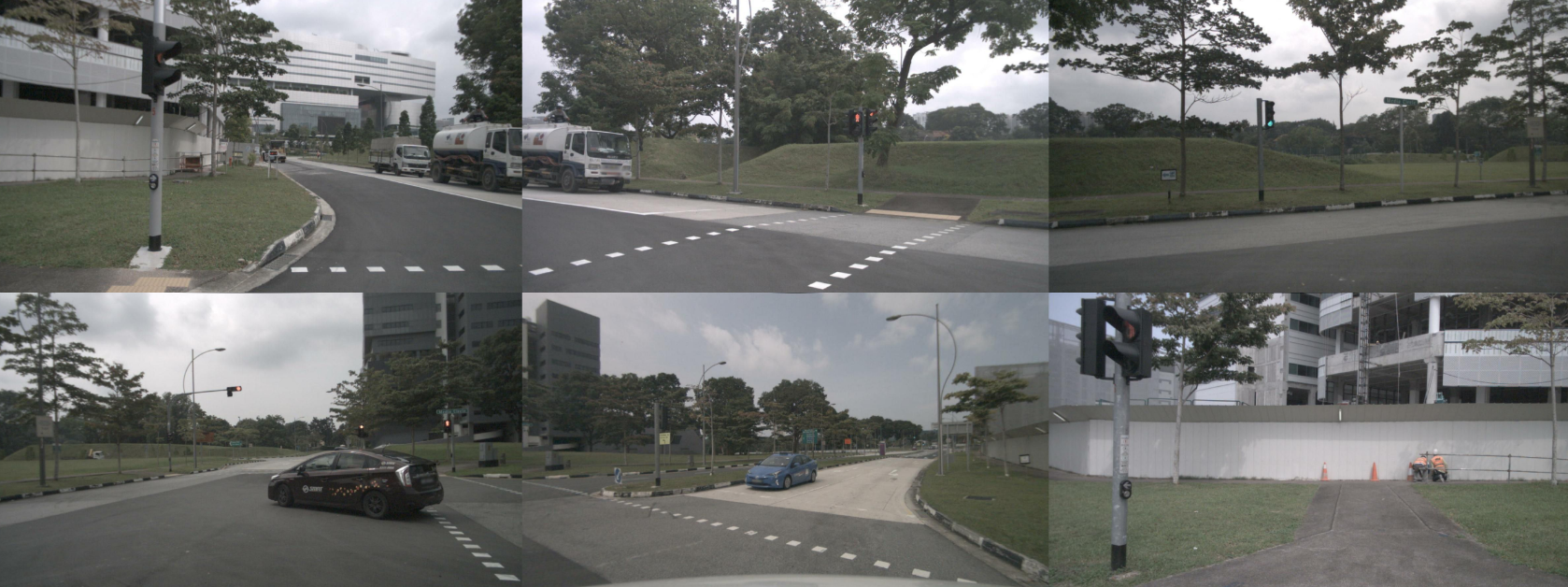}} & \\
\VarSty{ {\bf Prompt 1: Front Caption} } & \\
The front scene depicts a street intersection with several notable elements. On the left side of the front scene, there is a modern building with a white facade and large windows. In front of the building, there is a traffic light displaying a red signal. A parked vehicle is visible near the building, and there is a grassy area with a few trees and a sidewalk. In the center of the front scene...\textless omitted\textgreater & \\

\VarSty{ {\bf Prompt 2: Back Caption} } & \\
The back scene depicts a street intersection with various elements captured from different angles. On the left side of the back scene, there is a traffic light and a construction area with a fence and some equipment, including a small vehicle or machinery. Two traffic cones are placed near the fence, indicating a possible work zone or restricted area. The background features a multi-story building with visible structural elements...\textless omitted\textgreater & \\

\VarSty{ {\bf Prompt 3: Rewriting Caption} } & \\
The scene depicts a street intersection with various vehicles and elements. There are multiple vehicles, including trucks and cars, some of which are moving and others parked. Traffic cones are present in some parts of the scene, indicating possible construction or restricted areas. The area is surrounded by buildings and greenery, with traffic lights controlling the flow of vehicles. There are no visible pedestrians sitting or lying down. The setting appears to be an urban environment with infrastructure for traffic management. & \\

\hrulefill & \\
\VarSty{ {\bf Prompt 4: 3D Occupancy and Flow prediction } } & \\
Your task is to predict the 3D occupancy of the scene. Assume you are located at the point $(0, 0, 0)$. The scene area around you (in front, behind, left, and right) is divided into a $200\times200$ grid, with the bottom-left corner at $(-100, -100)$ and the top-right corner at $(100, 100)$. The height region is divided into 16 bins. We use $<OCC>(x, y, z)</OCC>$ to represent the point at location $(x, y)$ with a height of z. Assume you are located at the point $<OCC>(100, 100, 0)</OCC>$. Answer the below question. \\
Is the position \{position\} occupied?\\
What object is occupying position $<OCC>(x, y, z)</OCC>$? If there is an object, please provide its name and predict the velocity; otherwise, answer `free'. & \\

\VarSty{ {\bf Prompt 5: Action Prediction } } & \\
What is the safe action of the ego car? & \\

\VarSty{ {\bf Prompt 6: Trajectory Prediction } } & \\
Predict the future 6-frame trajectory of the ego car in the last. & \\

\end{tabular}
}
\end{tcolorbox}
\vspace{-10pt}
\label{tab:data}
\end{minipage}
\end{table*}

\noindent\textbf{Different Balancing Weights.}
In {\ourMethod}, we adopt multi-task learning to train our model in an end-to-end manner. Here, we conduct experiments to investigate the effect of different loss balancing weights on overall performance. Through preliminary analysis, we observe that the 3D occupancy and occupancy flow losses dominate the total loss function, accounting for over 60\% of the combined loss magnitude. To mitigate potential optimization imbalances, we systematically reduce their weights from the default value of 1.0~(\textit{III}) to 0.5~(\textit{II}) and 0.2~(\textit{I}) for both 3D occupancy and occupancy flow tasks.
The corresponding results are presented in Table~\ref{tab:ab_weight}. We find that higher occupancy and flow weights yield improved performance on 3D occupancy and occupancy flow, but result in slight degradation in planning accuracy~(L2 error) and text understanding in the QA task. Therefore, setting proper balancing weights in multi-task learning remains challenging. For simplicity, we adopt the default weight of 1.0 for both 3D occupancy and flow losses in our final implementation.

\noindent\textbf{Learned Weights of Different Hidden States.} 
For the MLLM~(\textit{i.e.}, Qwen2.5-0.5B/3B model), we investigate \textit{what is the importance weight of each hidden state of MLLMs in {\ourMethod}?} To answer this question, we conduct an analysis of the learned importance weights of all hidden states in the MLLM, as shown in Table~\ref{tab:layer_weights}. Specifically, we replace the default setting of using only the last hidden state with a weighted combination $h = \sum_{i=0}^l F_{i}^h \cdot w_i$, where $F_i^h \in \mathbb{R}^{N\times C}$ represents the features of the $i$-th hidden state, $w_i \in \mathbb{R}^{1\times1}$ is the corresponding learnable importance weight, and $l$ denotes the total number of hidden states in the MLLM. Here, the 0.5B~(or 3B) MLLM contains 24~(or 36) transformer layers, and when including the input embeddings, we obtain a total of 25~(or 37) hidden states with corresponding learnable weights. As shown in Table~\ref{tab:layer_weights} and Table~\ref{tab:3blayer_weights}, 
we observe an overall trend where deeper layers tend to receive larger weights, indicating that higher-level features extracted by deeper transformer layers are more crucial for the effectiveness of our {\ourMethod}.

\section{Details of Coarse-grained Spatial Understanding}
\label{sec:details_cos}

Table~\ref{tab:data} presents a variety of prompts generated by our multi-stage data engine. The data engine first produces individual captions for both front and back views, which are subsequently merged and refined into a coherent final caption. The corresponding prompts are shown as ``{\textcolor{black}{\bf Prompt 1-3}}" in Table~\ref{tab:data}. To enhance {\ourMethod}'s comprehensive spatial understanding capabilities, we design instruction question-answering~(QA) pairs covering four core tasks: 3D occupancy prediction, flow prediction, action prediction, and trajectory prediction. For 3D occupancy and flow prediction tasks, we generate questions about the occupancy status, category, and velocity of given 3D locations using occupancy and flow ground truth. The corresponding prompts are as shown in ``{\textcolor{black}{\bf Prompt 4: Occupancy and Flow Prediction}}" in Table~\ref{tab:data}. For action prediction, we design prompts to predict  subsequent driving actions, as demonstrated in ``{\textcolor{black}{\bf Prompt 5: Action Prediction}}" in Table~\ref{tab:data}. For trajectory prediction, we create prompts that guide the model to predict future ego-vehicle trajectories, as shown in ``{\textcolor{black}{\bf Prompt 6: Trajectory Prediction}}" in Table~\ref{tab:data}.

\section{Details of Fine-grained Spatial Learning}
\label{sec:details_fine}

We provide details of task-specific heads, including 3D occupancy, flow, and action diffusion heads for fine-grained spatial learning.

\noindent\textbf{Details of 3D Occupancy Head.}
For the 3D occupancy head, we mainly follow the previous superior method FlashOcc~\cite{yu2023flashocc}. Specifically, given the spatial feature map $F_{out} \in \mathbb{R}^{H\times W \times C}$ (refer to the main paper), we perform a reshape operation along the channel dimension to transform $F_{out}$ from a shape of $H\times W \times C$ to $H\times W \times Z \times C'$, where $C'$ and $Z$ represent the channel dimension and depth dimension of the features, respectively. Then, we use an MLP to predict the category of 3D occupancy. For loss functions, we adopt the same losses as in~\cite{yu2023flashocc}, including the focal loss $\mathcal{L}_{focal}$, geometric loss $\mathcal{L}_{geo}$, semantic loss $\mathcal{L}_{sem}$, and Lovász loss $\mathcal{L}_{lovasz}$.

\noindent\textbf{Details of Occupancy Flow Head.}
We use an additional MLP to predict the velocity of 3D occupancy based on the occupancy head. To address the imbalance in the distribution of static and moving objects, we employ L1 loss and apply distinct loss weights to the flow prediction task for static and dynamic occupancy, respectively. Specifically, we assign a weight of 1.0 to each dynamic occupancy flow for computing the flow loss, while applying a smaller weight of 0.01 to static occupancy flow during training.

\noindent\textbf{Details of Action Diffusion Head.}
We refer to previous action diffusion methods~\cite{chi2025diffusion,diffusiondrive} to implement a simple action diffusion head that predicts trajectories based on the denoising procedure. On nuScenes~\cite{caesar2020nuscenes}, DiffusionDrive~\cite{diffusiondrive} applies SparseDrive~\cite{sun2024sparsedrive} to achieve plan query initialization for better trajectory planning results. In contrast, our approach directly generates trajectories  without requiring SparseDrive for plan query initialization. For the loss of action diffusion head, we employ L1 loss for the action diffusion head during training.



\section{More Visualization}
\label{sec:more_vis}

In this section, we provide additional visualizations to demonstrate the coarse-grained and fine-grained understanding capabilities of {\ourMethod}. 
As shown in Table~\ref{tab:stationary}, {\ourMethod} generates accurate 3D occupancy and flow predictions in complex congested environments, and successfully implements appropriate actions for stationary waiting scenarios.
Furthermore, we visualize a straight-ahead driving scene in Table~\ref{tab:go}, where {\ourMethod} demonstrates high consistency between coarse-grained and fine-grained understanding. We also present a challenging nighttime turning scenario in Table~\ref{tab:turn}, which is typically difficult for VLA models. In this scene, leveraging multi-modal information, {\ourMethod} accurately describes the environment despite the very dark imagery. Therefore, our {\ourMethod} delivers accurate and reasonable results in this challenging low-light condition, demonstrating its robustness and effectiveness.

\begin{table*}[t!]
\centering
\vspace{-2mm}
\begin{minipage}{0.85\textwidth}  
\centering
\begin{tcolorbox} 
\centering
\scalebox{0.85}{
\begin{tabular}{p{1.1\textwidth} c}  
\centering{\includegraphics[height=4.3cm]{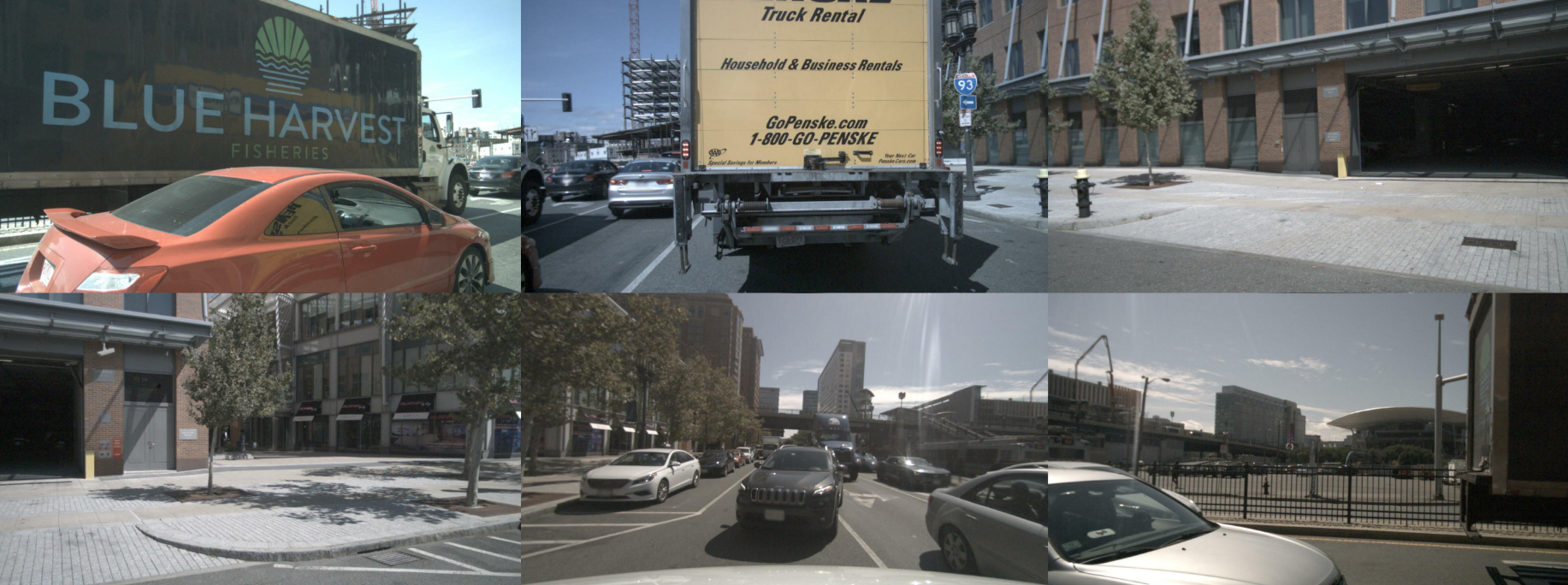}} & \\
\hrulefill & \\
\VarGSty{ {\bf Coarse-grained Perception, Prediction, and Planning} } & \\
\VarSty{ {\bf Task 1: Caption} } & \\
The scene depicts a quiet residential street lined with houses and lush greenery. The street is mostly empty, with a few parked vehicles visible in some parts. There are trees providing shade along the sidewalks, and the houses have gated entrances and fences. A bicycle is parked on the sidewalk near one of the houses. The overall atmosphere is calm and suburban, with no pedestrians or moving vehicles present. & \\

\VarSty{ {\bf Task 2: Occupancy and Flow prediction} } & \\
Question: What can be seen at position $<OCC>(83, 99, 5)</OCC>$? If there is an object, please provide its name and predict the velocity; otherwise, answer `free' \\
Answer: \{label: car\}, \{vx: 0.1, vy: 0.06\} & \\

\VarSty{ {\bf Task 3: Action decision} } & \\
Question: What is the safe action of the ego car?\\
Answer: Go straight. & \\

\VarSty{ {\bf Task 4: Trajectory prediction} } & \\
Question: Predict the future 6-frame trajectory of the ego car in the last. \\
Answer: Future 6-frame trajectory: [(0.12, 0.00), (0.05, 0.00), (-0.01, 0.00), (0.00, 0.00), (0.00, 0.00), (0.00, 0.00)]. & \\

\hrulefill & \\
\VarGSty{ {\bf Fine-grained Perception, Prediction, and Planning} } & \\
\centering{\includegraphics[height=4.3cm]{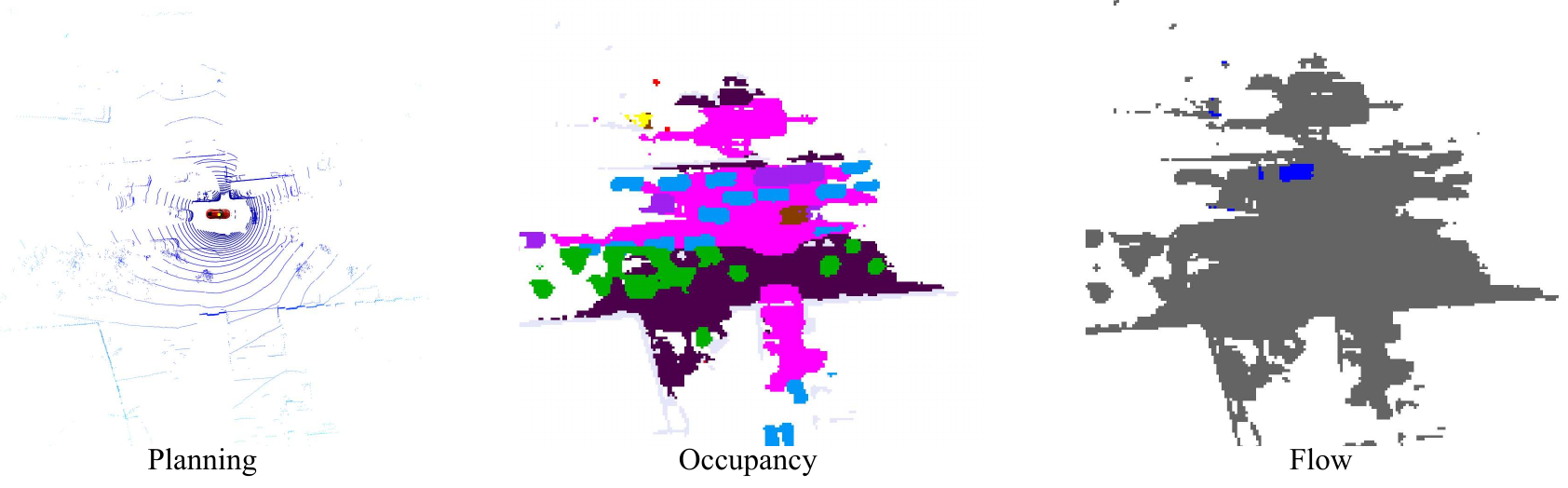}} & \\
\end{tabular}
}
\end{tcolorbox}
\vspace{-2mm}
\caption{The visualization of remaining stationary with coarse-grained and fine-grained results.}
\label{tab:stationary}
\end{minipage}
\end{table*}

\begin{table*}[t!]
\centering
\vspace{-2mm}
\begin{minipage}{0.85\textwidth}  
\centering
\begin{tcolorbox} 
\centering
\scalebox{0.85}{
\begin{tabular}{p{1.1\textwidth} c}  
\centering{\includegraphics[height=4.3cm]{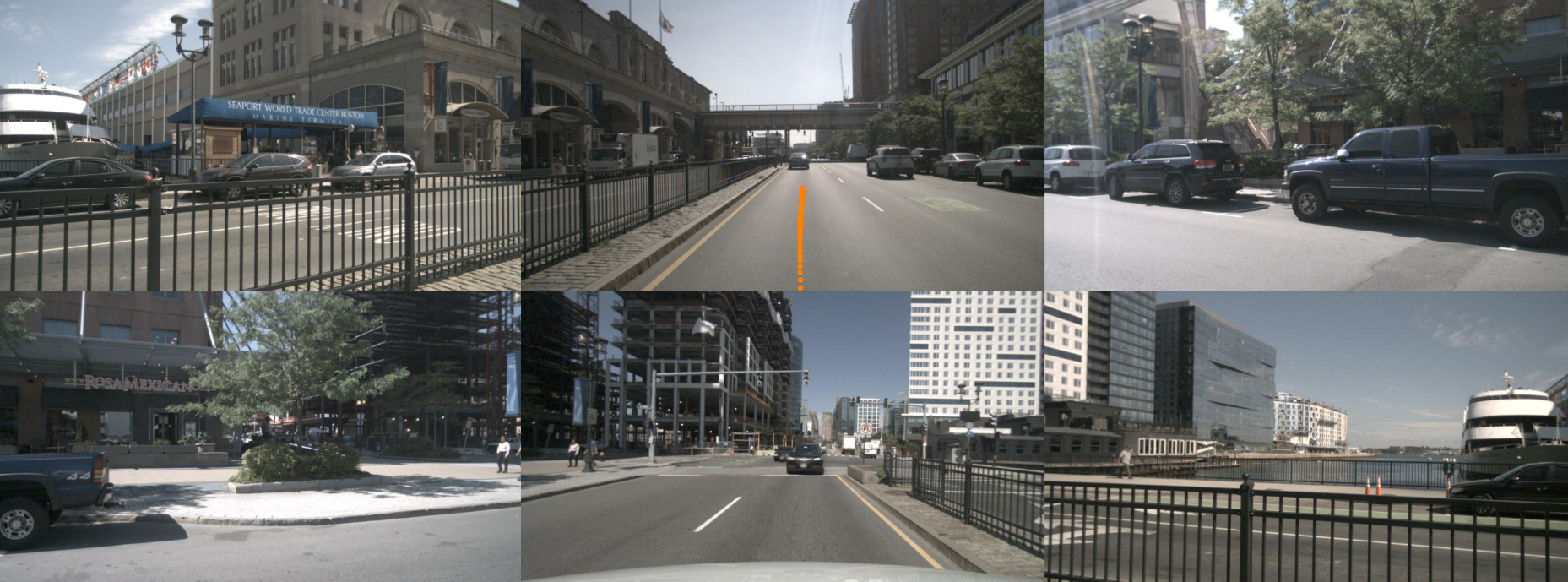}} & \\
\hrulefill & \\
\VarGSty{ {\bf Coarse-grained Perception, Prediction, and Planning} } & \\
\VarSty{ {\bf Task 1: Caption} } & \\
The scene depicts a busy urban street with several vehicles, both parked and moving. There are multiple cars, some stopped at intersections and others in motion. Pedestrians are visible walking along the sidewalks. Construction cranes and buildings under construction are present in the background, indicating ongoing development. Traffic cones are placed on the road, likely for directing traffic or marking construction areas. The environment appears to be a bustling city area with a mix of modern architecture and active traffic. & \\

\VarSty{ {\bf Task 2: Occupancy and Flow prediction} } & \\
Question: What can be seen at position $<OCC>(159, 78, 4)</OCC>$? If there is an object, please provide its name and predict the velocity; otherwise, answer `free' \\
Answer: \{label: car\}, \{vx: 0.1, vy: -0.07\} & \\

\VarSty{ {\bf Task 3: Action decision} } & \\
Question: What is the safe action of the ego car? \\
Answer: Go straight. & \\

\VarSty{ {\bf Task 4: Trajectory prediction} } & \\
Question: Predict the future 6-frame trajectory of the ego car in the last. \\
Answer: Future 6-frame trajectory: [(3.42, -0.03), (3.52, -0.02), (3.59, -0.04), (3.72, -0.07), (3.87, -0.10), (3.88, -0.09)]. & \\

\hrulefill & \\
\VarGSty{ {\bf Fine-grained Perception, Prediction, and Planning} } & \\
\centering{\includegraphics[height=4.3cm]{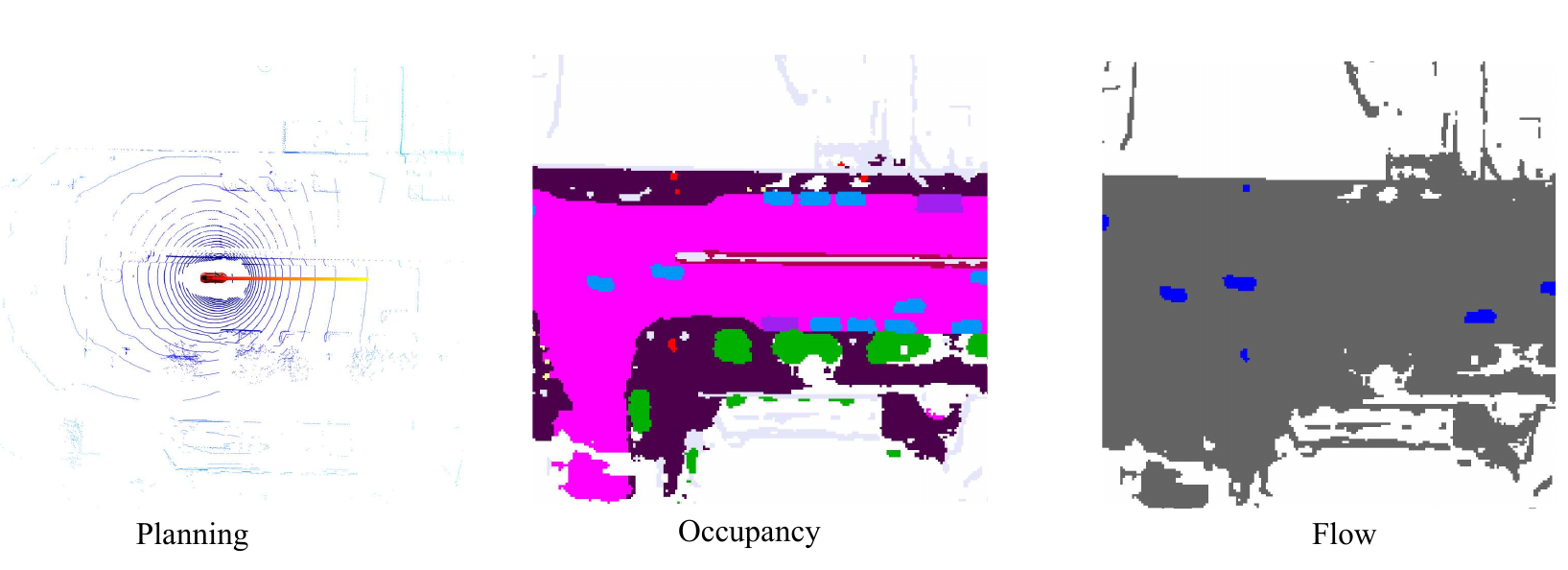}} & \\
\end{tabular}
}
\end{tcolorbox}
\vspace{-2mm}
\caption{The visualization of going straight with coarse-grained and fine-grained results.}
\label{tab:go}
\end{minipage}
\end{table*}

\begin{table*}[t!]
\centering
\vspace{-2mm}
\begin{minipage}{0.85\textwidth}  
\centering
\begin{tcolorbox} 
\centering
\scalebox{0.85}{
\begin{tabular}{p{1.1\textwidth} c}  
\centering{\includegraphics[height=4.3cm]{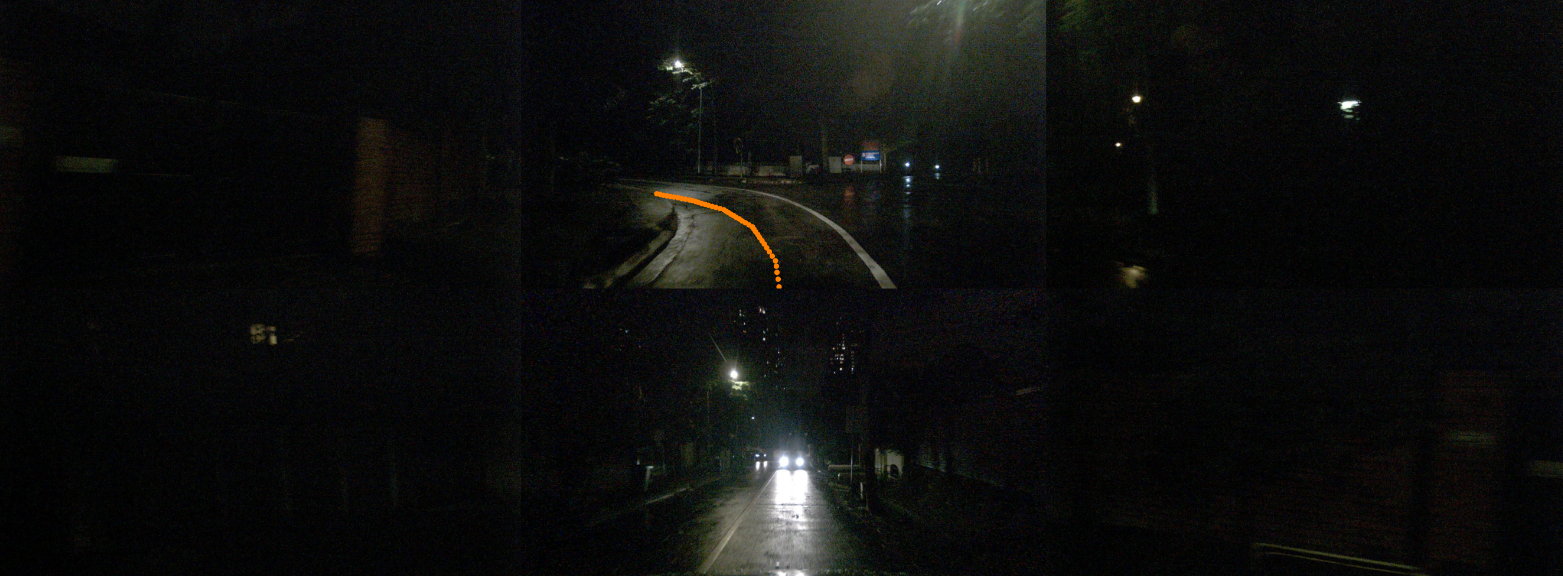}} & \\
\hrulefill & \\
\VarGSty{ {\bf Coarse-grained Perception, Prediction, and Planning} } & \\
\VarSty{ {\bf Task 1: Caption} } & \\
The scene depicts a quiet residential street lined with houses and lush greenery. The road is paved and has a few parked vehicles, including a white van and a car. There are trees and plants along the sidewalks, and some trash bins are visible near the houses. The area appears calm and suburban, with no pedestrians or moving vehicles in sight. & \\

\VarSty{ {\bf Task 2: Occupancy and Flow prediction} } & \\
Question: What can be seen at position $<OCC>(173, 102, 8)</OCC>$? If there is an object, please provide its name and predict the velocity; otherwise, answer `free' \\
Answer: \{label: vegetation\}, \{vx: 0.0, vy: 0.0\} & \\

\VarSty{ {\bf Task 3: Action decision} } & \\
Question: What is the safe action of the ego car?\\
Answer: Turn left. & \\

\VarSty{ {\bf Task 4: Trajectory prediction} } & \\
Question: Predict the future 6-frame trajectory of the ego car in the last. \\
Answer: Future 6-frame trajectory: [(3.55, -0.02), (3.34, -0.25), (3.12, -0.68), (2.98, -1.17), (2.69, -1.74), (2.47, -2.18)]. & \\

\hrulefill & \\
\VarGSty{ {\bf Fine-grained Perception, Prediction, and Planning} } & \\
\centering{\includegraphics[height=4.3cm]{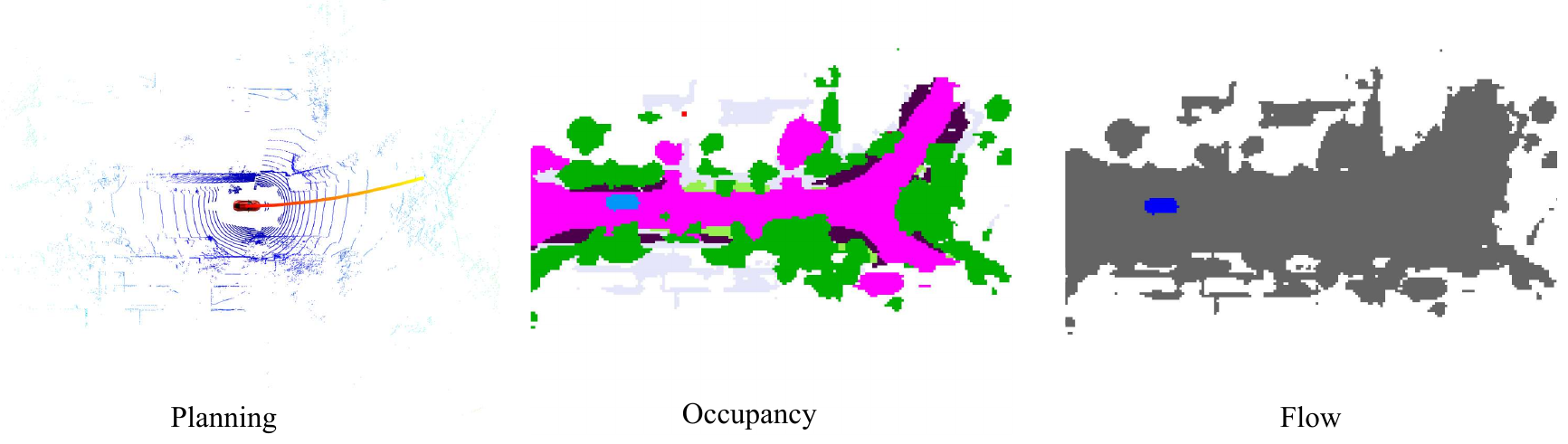}} & \\
\end{tabular}
}
\end{tcolorbox}
\vspace{-2mm}
\caption{The visualization of turning with coarse-grained and fine-grained results.}
\label{tab:turn}
\end{minipage}
\end{table*}

\end{document}